\documentclass[lettersize,journal]{IEEEtran}
\usepackage{amsmath,amsfonts}
\usepackage{algorithmic}
\usepackage{algorithm}
\usepackage{array}
\usepackage[caption=false,font=normalsize,labelfont=sf,textfont=sf]{subfig}
\usepackage{textcomp}
\usepackage{stfloats}
\usepackage{url}
\usepackage{verbatim}
\usepackage{graphicx}
\usepackage{cite}
\usepackage{bm}
\usepackage{pifont}
\begin{document}

\title{ILBiT: Imitation Learning for Robot Using Position and Torque Information based on Bilateral Control with Transformer}

\author{
\IEEEauthorblockN{Masato Kobayashi$^{1*}$, Thanpimon Buamanee$^{2}$, Yuki Uranishi$^{1}$, Haruo Takemura$^{1}$}
\thanks{
$1$ Cybermedia Center, Osaka University, Toyonaka, Osaka, 560-0043 Japan,
$2$ School of Engineering Science, Osaka University, Toyonaka, Osaka 560-0043, Japan,
$^{*}$Email:kobayashi.masato.cmc@osaka-u.ac.jp
}
}



\maketitle

\begin{abstract}
Autonomous manipulation in robot arms is a complex and evolving field of study in robotics.
This paper introduces an innovative approach to this challenge by focusing on imitation learning (IL).
Unlike traditional imitation methods, our approach uses IL based on bilateral control, allowing for more precise and adaptable robot movements.
The conventional IL based on bilateral control method have relied on Long Short-Term Memory (LSTM) networks.
In this paper, we present the IL for robot using position and torque information based on Bilateral control with Transformer (ILBiT).
This proposed method employs the Transformer model, known for its robust performance in handling diverse datasets and its capability to surpass LSTM's limitations, especially in tasks requiring detailed force adjustments.
A standout feature of ILBiT is its high-frequency operation at 100 Hz, which significantly improves the system's adaptability and response to varying environments and objects of different hardness levels.
The effectiveness of the Transformer-based ILBiT method can be seen through comprehensive real-world experiments. 
\end{abstract}


\section{Introduction}
\label{sec:introduction}


In Imitation learning (IL), robotic arms can learn manipulative tasks by mimicking the actions demonstrated by human experts. One mainstream approach within IL is Behavioral Cloning (BC), which involves learning a function that maps observations to actions from an expert's demonstrations using supervised learning\cite{robobc2020ly, robobc2023shukla}.
Another prominent approach in IL is Inverse Reinforcement Learning (IRL), which aims to infer a reward function from the demonstrations\cite{irl2000ng, irl2004abbeel, irl2021arora}. Contrary to IRL, BC does not require additional trials for training manipulation. This paper focuses on the IL approach utilizing BC, a technique founded on supervised learning.

In the field of IL, the method of collecting robot data is a crucial factor\cite{lfd2009argall, lfd2022mukherjee}.
Demonstrations for IL are collected through various methods such as video observation\cite{vo2014hayes, vo2018liu}, human-guided teaching\cite{hc2019wang}, domain transformation\cite{dt2023kobayashi, dt2020zhang}, and teleoperation\cite{topvr2018Zhang, topvr2023Khansari, topsma2018Ajay, topdirect2021Johns, topdirect2022Valassakis, topvideo2021Xiong}. Learning from video observation involves imitating tasks based on information extracted from video footage\cite{vo2014hayes, vo2018liu}. In domain transformation, tasks are performed by both humans and robots, with techniques available to convert human-related information into robot-compatible formats\cite{dt2023kobayashi}. This paper primarily focuses on data collection through teleoperation.
This is often facilitated by employing various devices, including virtual reality (VR) headsets and hand tracking systems \cite{topvr2018Zhang, topvr2023Khansari}, as well as smartphones\cite{topsma2018Ajay}, direct control systems \cite{topdirect2021Johns, topdirect2022Valassakis}, and human-video interaction techniques \cite{topvideo2021Xiong}.
These innovations indicate a trend towards more advanced and user-friendly data collection methods in robot control, resulting in increasingly flexible ways to utilize robots.

The data collected through these remote operations are used for IL.
Kim et al. have implemented IL using a transformer model that takes as inputs the operator's gaze, images, and the position and orientation information of the robot's end effector, with the position and orientation of the robot's end effector as the output, operating at 10Hz\cite{imip2021Kim}.
Meanwhile, Brohan et al. have implemented Robotics Transformer (RT-1). RT-1 is conducting IL with images and text as the input and the action of the robot's end effector, base, and mode as the output at 3Hz\cite{imip2023rt1}.
However, IL, particularly when relying on position control, a common strategy in robotics, faces specific challenges \cite{imip2021Jang, imip2021Kim, imip2023rt1, imip2022Hamano}. 
The first point to note is that many robotic systems operate at relatively low action inference frequencies, often within the range of 1-10 Hz, which results in slower movements of the robots. This makes it challenging for them to move at speeds equivalent to humans. Another aspect is position control, which involves instructing the robot to achieve and maintain specific positions based on learned human actions. This method can struggle with environmental changes or managing objects of varying hardness and sizes. While it remains a promising approach in robotic manipulation, these challenges in adaptability suggest the need for further refinement in IL techniques.

\begin{figure}[t]
  \begin{center}
    \scalebox{0.35}{
        \includegraphics{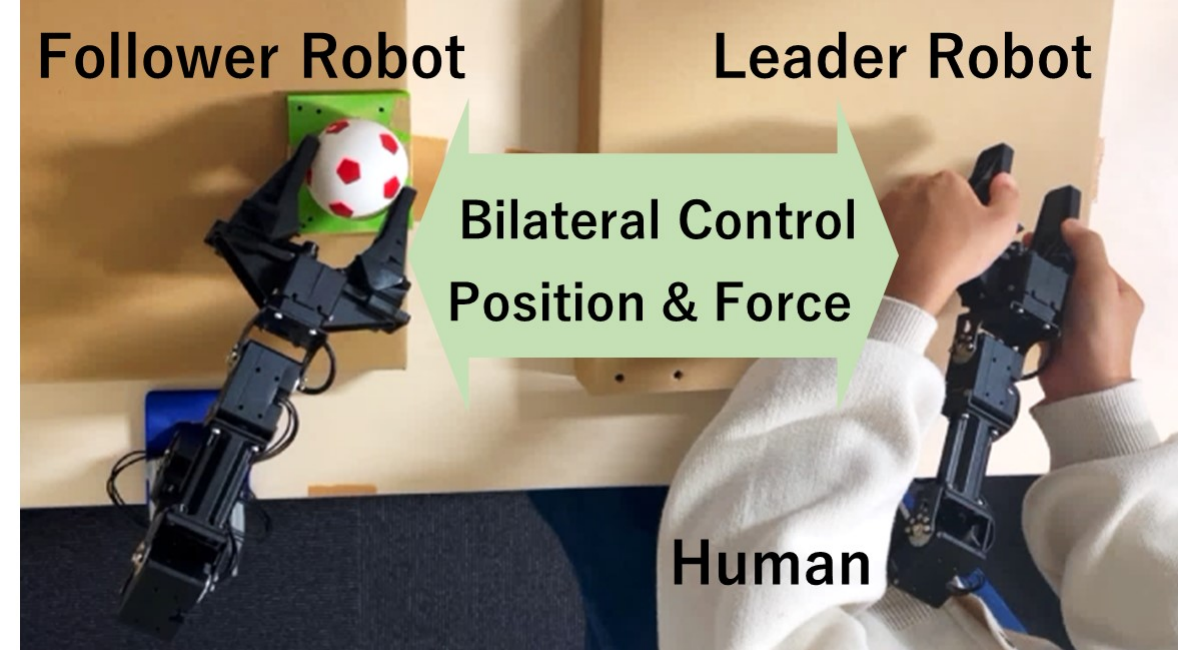}}    
  \caption{Image of Bilateral Control}
  \label{fig:ibi}
\end{center}
\end{figure}
To address these limitations, we explore the potential of IL based on bilateral control, a hybrid technique that combines aspects of both position and force control \cite{IMIB2018adachi, IMIB2019fujimoto, IMIB2020sasagawa, IMIB2021sasagawa, IMIB2022sugiura, IMIB2022sakaino, IMIB2022hayashi, IMIB2023yamane}.
As shown in Fig.~\ref{fig:ibi}, the operator controls the leader robot through bilateral control, and the follower robot moves in synchronization with its motion. Bilateral control is realized through the tracking of positions and the principle of action-reaction, allowing the operator to feel the force and tactile information of the follower robot.
Traditional remote operation methods for data collection utilized the robot's response values as learning data. However, using response values does not consider factors like control delays, which can hinder the generation of robot actions. A distinctive feature of IL based on bilateral control is the use of control command values, rather than control response values, as learning data.
The learning data also involves using the robot's joint angle, velocity, and torques, with the output also employing these parameters. Furthermore, because operators can sense the hardness or softness of objects during data collection in remote operation utilizing bilateral control, it proves to be a better method for data collection. Additionally, robotic systems trained with these data usually operate within a model inference range of 25-50 Hz, leading to faster movements of the robot.

Bilateral control in IL has shown promise in tasks requiring precise force modulation, enabling the robot to learn rapid, human-like motions. Conventionally, this approach has been implemented using Long Short-Term Memory (LSTM) networks, known for their effectiveness in learning and predicting time series data\cite{lstm}.
However, LSTM struggles with extremely high diversity in datasets. In response, this paper introduces a novel approach: IL based on bilateral control utilizing Transformer \cite{TRANS2017vaswani}.
Transformer can manage longer time series data and a broader range of data more effectively than LSTM.
The distinction between the research in this paper and previous studies\cite{imip2021Kim, imip2023rt1} that utilized transformers for position control in IL lies in our focus on bilateral control with current (torque) control that handles both position and force information.
Our approach of employing transformers in bilateral control for IL, particularly with an emphasis on both position and force data through current control.
We validate the effectiveness of our proposed method, which we refer to as ILBiT, through experiments involving pick-and-place tasks with objects of varying rigidity.

The main contributions of this paper are twofold:
\begin{itemize}
\item This paper proposes a novel approach to IL based on bilateral control utilizing Transformer, hereafter referred to as ILBiT.
ILBiT is designed to dynamically adjust to varying environmental conditions and object hardness, overcoming the limitations of the position and LSTM based methods.
\item ILBiT leverages the advanced capabilities of Transformer to handle diverse datasets, which allows for more effective learning and prediction of time series data compared to LSTM-based methods.
\end{itemize}
Finally, the effectiveness of ILBiT has been validated through extensive real-world experiments involving pick-and-place tasks with objects of varying hardness.

This paper is organized into five sections including the current section.
Section II provides a comprehensive overview of the robot system.
Section III proposes an approach to IL for robot arms using position and torque information based on bilateral control with the transformer (ILBiT).
Section IV shows the results from our real-world experiments to highlight the effectiveness and utility of ILBiT.
Finally, section V provides conclusions.

\section{Robot System}
\subsection{Controller}
\begin{figure}[t]
  \begin{center}
    \scalebox{0.27}{
        \includegraphics{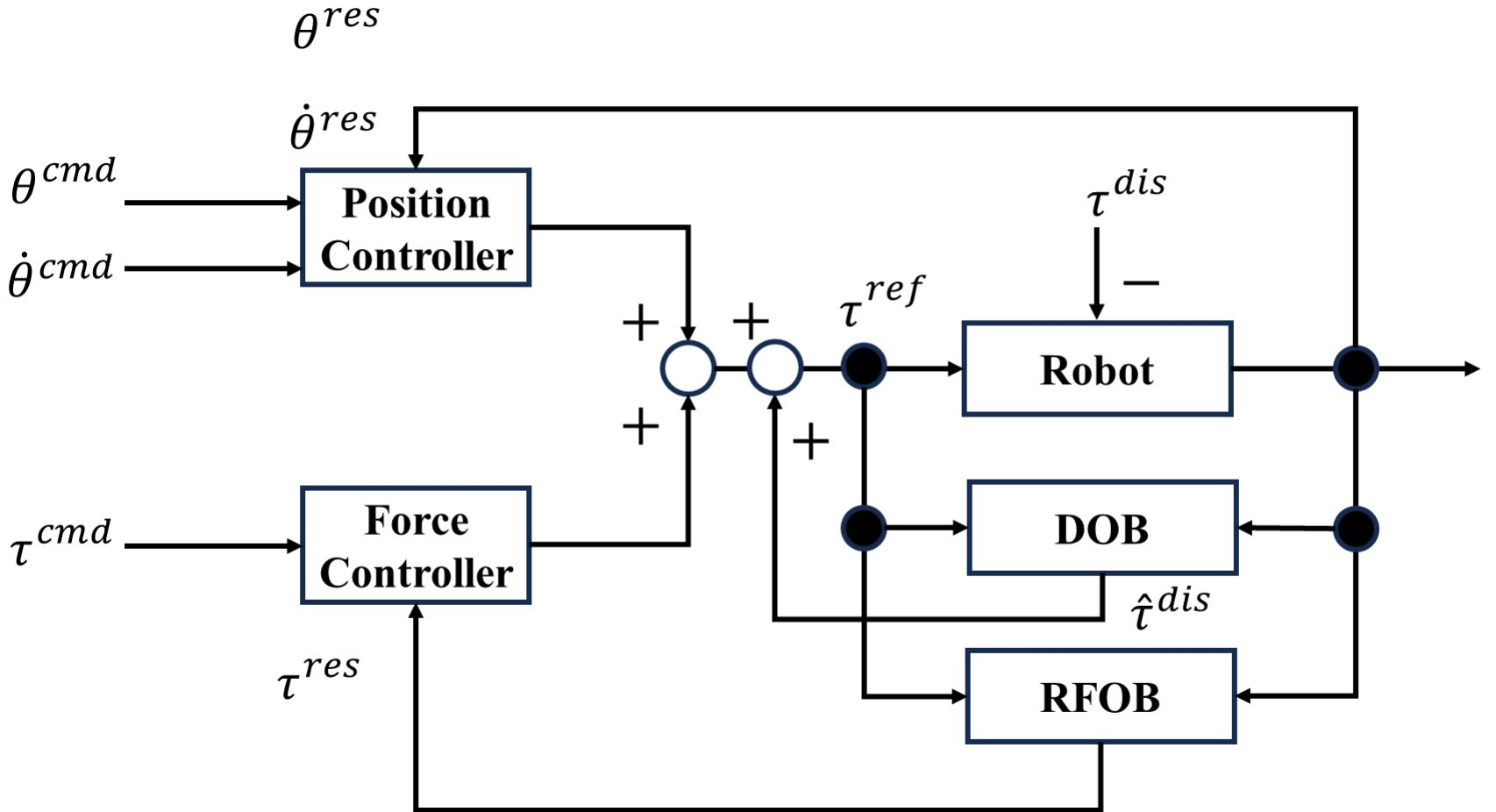}}    
  \caption{Block Diagram of Robot System}
  \label{fig:robot_system}
\end{center}
\end{figure}

The controller design adopted a control of position and force for each axis, as shown in Fig.~\ref{fig:robot_system}.
Angle information was obtained from encoders, and angular velocity was calculated by differentiating this information. The disturbance torque $\hat{\tau}^{dis}$ was calculated using a disturbance observer (DOB) \cite{DOB}, and the torque response value $\tau^{res}$ was estimated using a force reaction observer (RFOB) \cite{RFOB}.
\subsection{Bilateral Control}
\begin{figure}[t]
  \begin{center}
    \scalebox{0.3}{
        \includegraphics{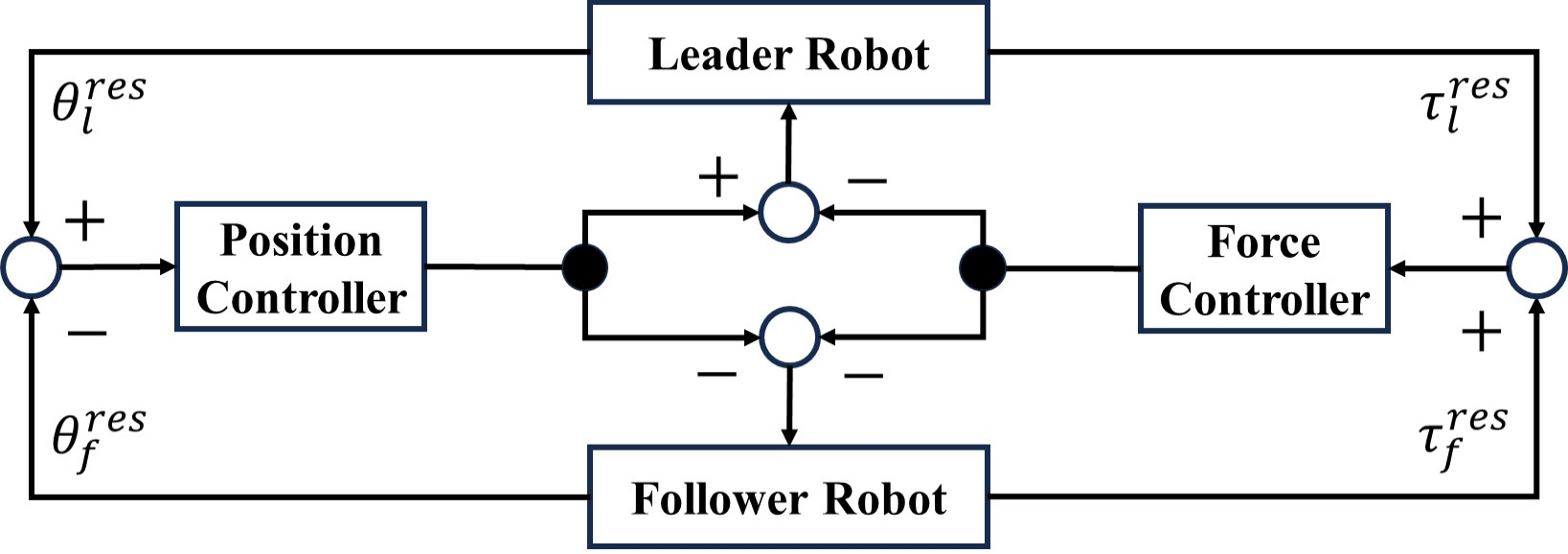}}    
  \caption{Block Diagram of Four-channel Bilateral Control}
  \label{fig:4ch}
\end{center}
\end{figure}

Fig.~\ref{fig:ibi} shows the bilateral control by OpenMANIPULATOR-X.
The fundamental principle of bilateral control is sharing position, force, or other information between the operator and the control target.
The control goals of the bilateral control are summarized as follows:
\begin{equation}
\theta_l - \theta_f = 0
\label{eq:position} 
\end{equation}
\begin{equation}
\tau_l + \tau_f = 0
\label{eq:force}
\end{equation}
where $\theta$ and $\tau$ represents the joint angle and torque.
The subscript $\bigcirc_l$ represents the leader system, and $\bigcirc_f$ represents the follower system.
This allows the operator to perform intuitive control over the control target.
Specifically, bilateral control is achieved by satisfying (\ref{eq:position}), representing position tracking between systems, and (\ref{eq:force}), representing the action-reaction relationship of forces.

As shown in Fig.~\ref{fig:4ch}, Four-channel bilateral control was used to collect human operation data for IL. 
This control scheme consists of a human-operated leader and a follower robot system that tracks the leader's movements. It simultaneously controls position and torque, setting the joint angle of the leader and follower as matching target values in position control. For torque, the torque signs applied to the leader and follower joints are inverted. Thus, the follower replicates the human's operational feel, while the leader reproduces the follower's reaction force.
As shown in Fig.~\ref{fig:robot_system} and \ref{fig:4ch}, torque reference value vectors of the controller that satisfies equations (1) and (2) are represented by the following equations:
\begin{equation}
\tau_{l}^{ref} = J(K_p + K_d s)(\theta_{l}^{res} - \theta_{f}^{res}) - K_f(\tau_{l}^{res} + \tau_{f}^{res})+\hat{\tau}^{dis}
\label{eq:leader} 
\end{equation}

\begin{equation}
\tau_{f}^{ref} = -J(K_p + K_d s)(\theta_{l}^{res} - \theta_{f}^{res}) - K_f(\tau_{l}^{res} + \tau_{f}^{res})+\hat{\tau}^{dis}
\label{eq:follow} 
\end{equation}

where $J$, $K_p$, $K_d$, $K_f$, and $s$ are the inertia and gain of position, velocity, force, and Laplace operator.
Bilateral control is realized through (\ref{eq:leader}) and (\ref{eq:follow}).

\subsection{Imitation Learning System Based on Bilateral Control}
\begin{figure}[t]
  \begin{center}
    \scalebox{0.25}{
        \includegraphics{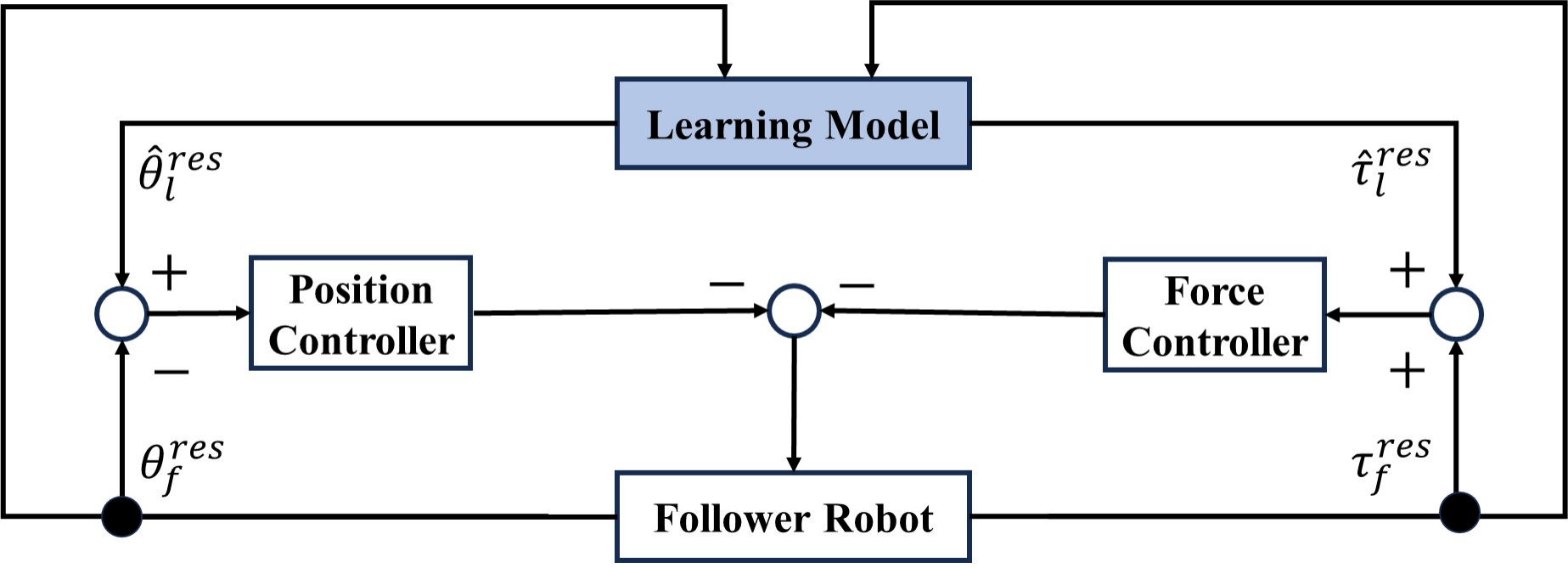}}    
  \caption{Block Diagram of Imitation Learning based on Bilateral Control}
  \label{fig:4ch_imi}
\end{center}
\end{figure}

\begin{figure}[t]
  \begin{center}
    \scalebox{0.25}{
        \includegraphics{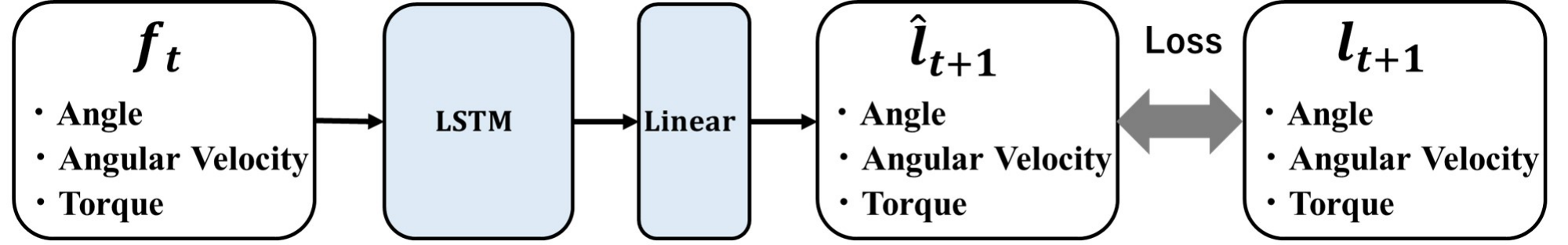}}    
  \caption{Image Diagram of LSTM Model (Conventional Method)}
  \label{fig:lstm}
\end{center}
\end{figure}
As shown in Fig.~\ref{fig:4ch} and \ref{fig:4ch_imi}, IL based on bilateral control \cite{IMIB2018adachi} is a method that gathers expert data using Four-channel bilateral control remote operation technology and imitates the human operational feel based on this information.
Traditional IL predicts the next operation from one operational result, causing motion delays due to robot control latency. However, with bilateral control-based IL, the next robot movement instruction is predicted from the actual results of the robot's actions, enabling rapid movements that account for control delays as shown in Fig.~\ref{fig:lstm}. In the diagram, "$\bm{f}$" represents the follower's joint angle, angular velocity, and torques, while "$\bm{l}$" represents the leader's corresponding vector information.
Here, the states of the follower and leader at the $t$-th time step are defined as follows:
\begin{equation}
\bm{l_t} = [\theta_l(t), \dot{\theta}_l(t), \tau_l(t)]
\end{equation}

\begin{equation}
\bm{f_t} = [\theta_f(t), \dot{\theta}_f(t), \tau_f(t)]
\end{equation}
where $\dot{\theta}$ represents the angular velocity.

\begin{figure*}[t]
\begin{center}
\scalebox{0.45}{
\includegraphics{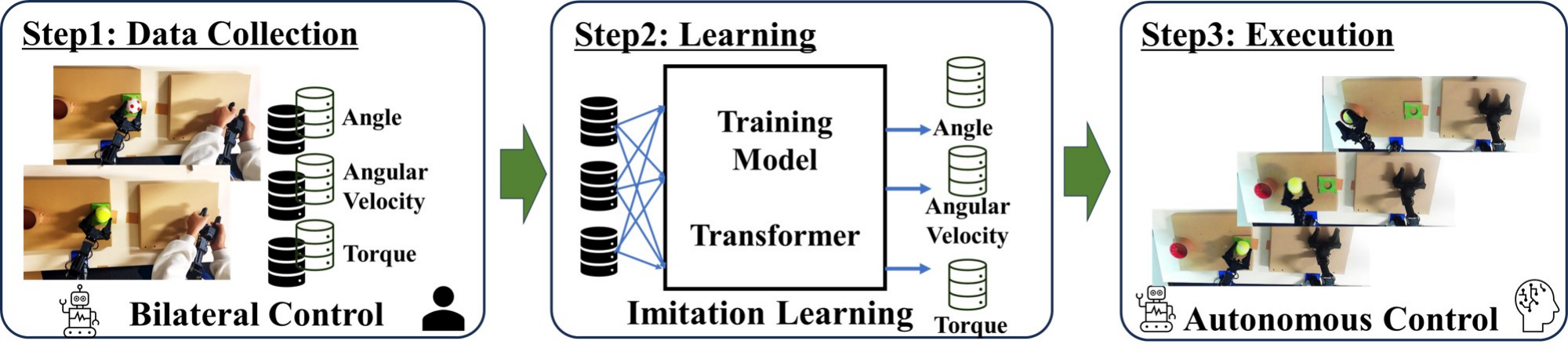}}
\end{center}
\caption{Overview of ILBiT}
\label{fig:overview_frame}
\end{figure*}

Bilateral control in IL has shown promise in tasks requiring precise force modulation, enabling the robot to learn rapid, human-like motions \cite{IMIB2018adachi, IMIB2019fujimoto, IMIB2020sasagawa, IMIB2021sasagawa, IMIB2022sugiura, IMIB2022sakaino, IMIB2022hayashi}.
Conventionally, this approach has been implemented using LSTM networks in the range of 25-50 Hz, known for their effectiveness in learning and predicting time series data.

\section{Imitation Learning based on Bilateral Control with Transformer (ILBiT)}
\subsection{Overall of ILBiT}
This section introduces a groundbreaking approach in IL named ILBiT, integrating bilateral control with Transformer technology.
This hybrid method combines the strengths of both position and force control, addressing the adaptability issues inherent in traditional methods.
Unlike LSTM networks which have been the standard in this domain, Transformer excels in managing diverse and extensive time series data, thus offering a substantial improvement over LSTM's capabilities.

Fig.~\ref{fig:overview_frame} presents an overview of ILBiT.
In this section, the implementation of ILBiT is structured into three distinct steps:

\begin{itemize}
  \item {\it Data Collection via Bilateral Control (Step 1)}\\
        As shown in Fig.~\ref{fig:overview_frame}, the first step involves the collection of demonstrative data for IL via bilateral control.

  \item {\it Transformer Model for Learning and Prediction (Step 2)}\\
        As illustrated in Fig.~\ref{fig:overview_frame}, Step 2 employs the Transformer model to process the collected data. This model excels in handling diverse datasets and is adept at learning complex patterns of human manipulation, enabling the system to predict and modulate forces accurately for effective robotic manipulation.

  \item {\it Execution through Trained Model (Step 3)}\\
      Fig.~\ref{fig:overview_frame} shows the system autonomously performing tasks with a robotic arm using the Transformer, showcasing adaptability and efficiency in real-world settings.

\end{itemize}

The details of each step are further elaborated in subsections III.B to III.D.

\subsection{Data Collection via Bilateral Control}
The initial phase of our approach is focused on the meticulous data collection through 'Bilateral Control'. In this stage, a robotic arm is configured to precisely follow and replicate the movements of a human user. This process is key to capturing the nuanced aspects of human motion, such as specific angle, velocity, and applied torques. The aim is to record a rich dataset that encompasses a wide range of human-led actions. These detailed recordings form the foundation for the robot's learning and subsequent autonomous performance. The data collected in this phase is crucial, as it represents the various human actions the robot will later aim to mimic through IL.

\subsection{Transformer Model for Learning and Prediction}
\begin{figure}[t]
\begin{center}
\scalebox{0.25}{
\includegraphics{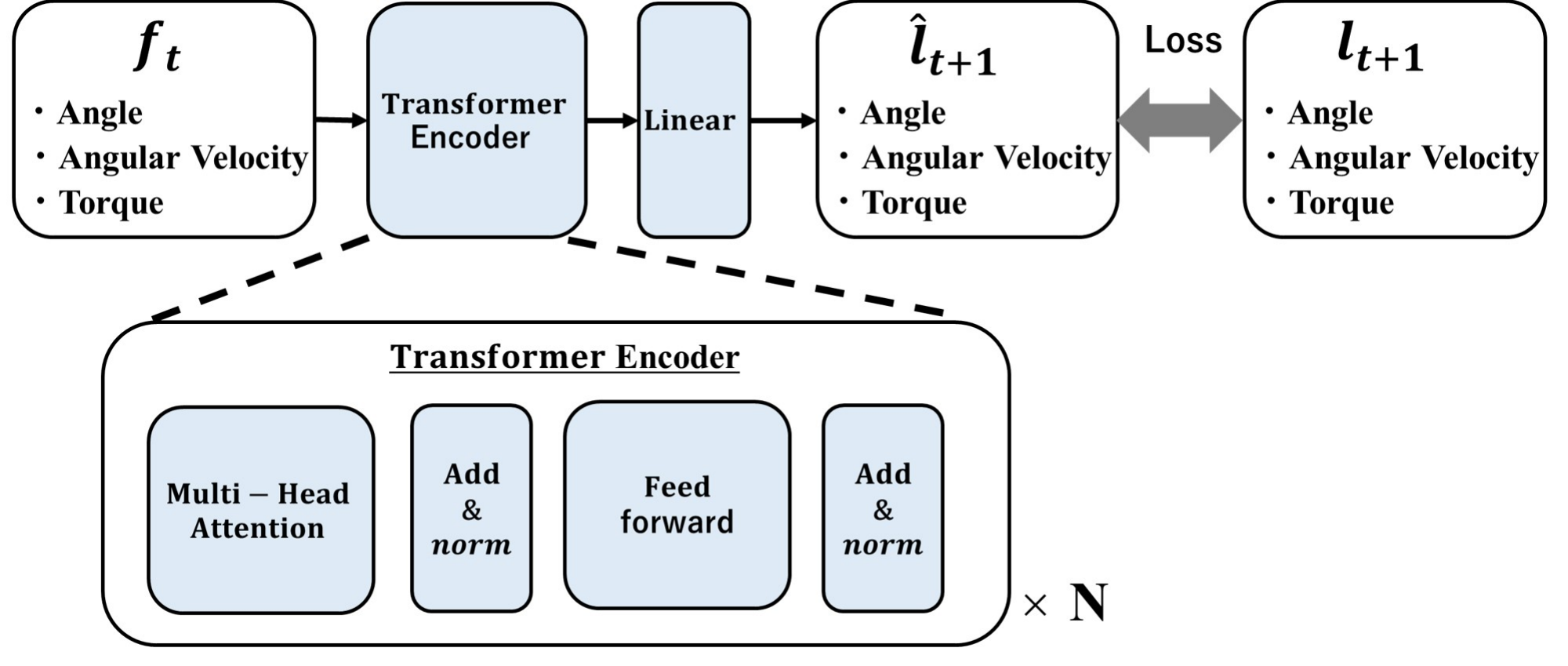}}
\caption{Image Diagram of the ILBiT Model (Proposed Method)}
\label{fig:TE}
\end{center}
\end{figure}
After accumulating a robust dataset, we transition to the core learning phase, where 'IL' comes into play. The collected data - encompassing parameters like angle, angular velocity, and torques - are fed into a learning model. 
This model is adept at handling sequential data, making it ideal for learning the temporal dynamics of human motion.
Unlike conventional LSTM methods, the Transformer architecture enables the system to process these sequences more effectively, capturing subtleties in the data.
This enhanced learning capability allows the robot to not only imitate the recorded human actions but also generalize these learnings to new, unseen scenarios.
It is through this advanced learning mechanism that the robot gains the ability to replicate human-like movements with higher precision and adaptability.

The architecture of the Transformer Encoder, central to our ILBiT method, is depicted in Fig.~\ref{fig:TE}.
The input to the Transformer, denoted as \( \bm{f_t} \), represents the follower's joint angle, angular velocity, and torques at time \( t \), capturing the dynamic state of the robotic arm.
The Transformer Encoder processes the input \( \bm{f_t} \) through a series of operations as follows:
\begin{enumerate}
    \item \textbf{Multi-Head Attention}: The input first passes through a multi-head attention layer, enabling the model to simultaneously attend to different segments of the input sequence, which is essential for capturing complex motion patterns.
    
    \item \textbf{Add \& Norm}: Subsequent to attention, the process includes residual connections—adding the original input to the output of the attention layer—and layer normalization, aiding in learning stability and gradient flow.
    
    \item \textbf{Feedforward}: The output from the 'Add \& Norm' layer is then processed by a feedforward neural network, facilitating the modeling of more intricate relationships within the data.
    
    \item \textbf{Add \& Norm}: This pattern of alternating between attention and feedforward layers, with residual connections and normalization, is repeated \( N \) times, with \( N \) denoting the number of encoder layers.
    
    \item \textbf{Linear Layer}: After the final normalization, the data is projected through a linear layer to produce the predicted leader's action vector \( \bm{\hat{l}_{t+1}} \) for the subsequent time step.
    
    \item \textbf{Loss}: The model's prediction \( \bm{\hat{l}_{t+1}} \) is compared against the actual leader's vector \( \bm{l_{t+1}} \) using a loss function, guiding the optimization of model parameters during training.
\end{enumerate}

Through iterative training and minimization of the loss, the Transformer Encoder is trained to predict the leader's actions with high fidelity, translating human expertise into robotic autonomy for manipulation tasks.

\subsection{Execution through Autonomous Control}
In the final phase, equipped with the knowledge gained from the IL process, the robot is capable of independently executing tasks. This phase is critical, as it showcases the real-world application of the robot's learned skills. For instance, the robot demonstrates its proficiency in pick-and-place tasks, a fundamental yet complex activity in manipulation. Operating at a frequency of 100Hz, the robot is not only able to perform these tasks with high accuracy but also adapt to different objects and environments. This level of autonomy highlights the effectiveness of our ILBiT approach in teaching robots to perform intricate tasks with human-like dexterity and adaptability in various settings.

\section{Experiment}
\subsection{Hardware}
\begin{figure}[t]
  \begin{center}
    \scalebox{0.23}{
        \includegraphics{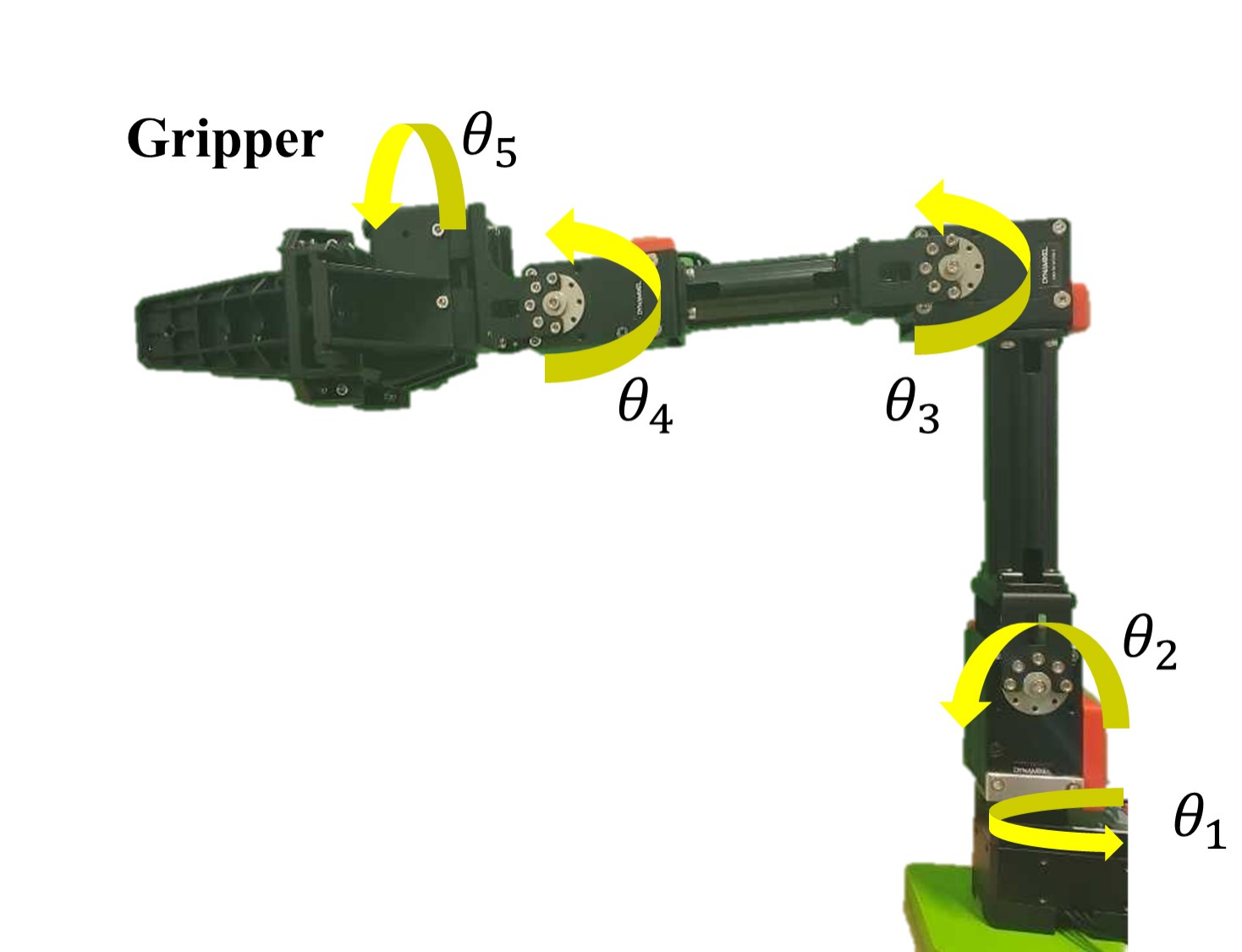}}    
  \caption{Definition of Robot's Joints (OpenMANIPULATOR-X)}
  \label{fig:OMX}
\end{center}
\end{figure}
In the conducted experiments, the OpenMANIPULATOR-X robotic arm, manufactured by ROBOTIS, was utilized as shown in Fig.~\ref{fig:OMX}.
This robotic arm is characterized by its four degrees of freedom, which facilitate its multidirectional movement.
Additionally, the gripper, a crucial component for object manipulation, possesses an individual degree of freedom.
Thus, the entire system exhibits a composite of five degrees of freedom. The actuation of these components is driven by DYNAMIXEL XM430-W350-T actuators, ensuring precise and reliable motion control. The control cycle period was set to 100 Hz.

\subsection{Environment Setup}
\begin{figure}[t]
  \begin{minipage}{0.49\hsize}
    \begin{center}
      \scalebox{0.15}{
        \includegraphics{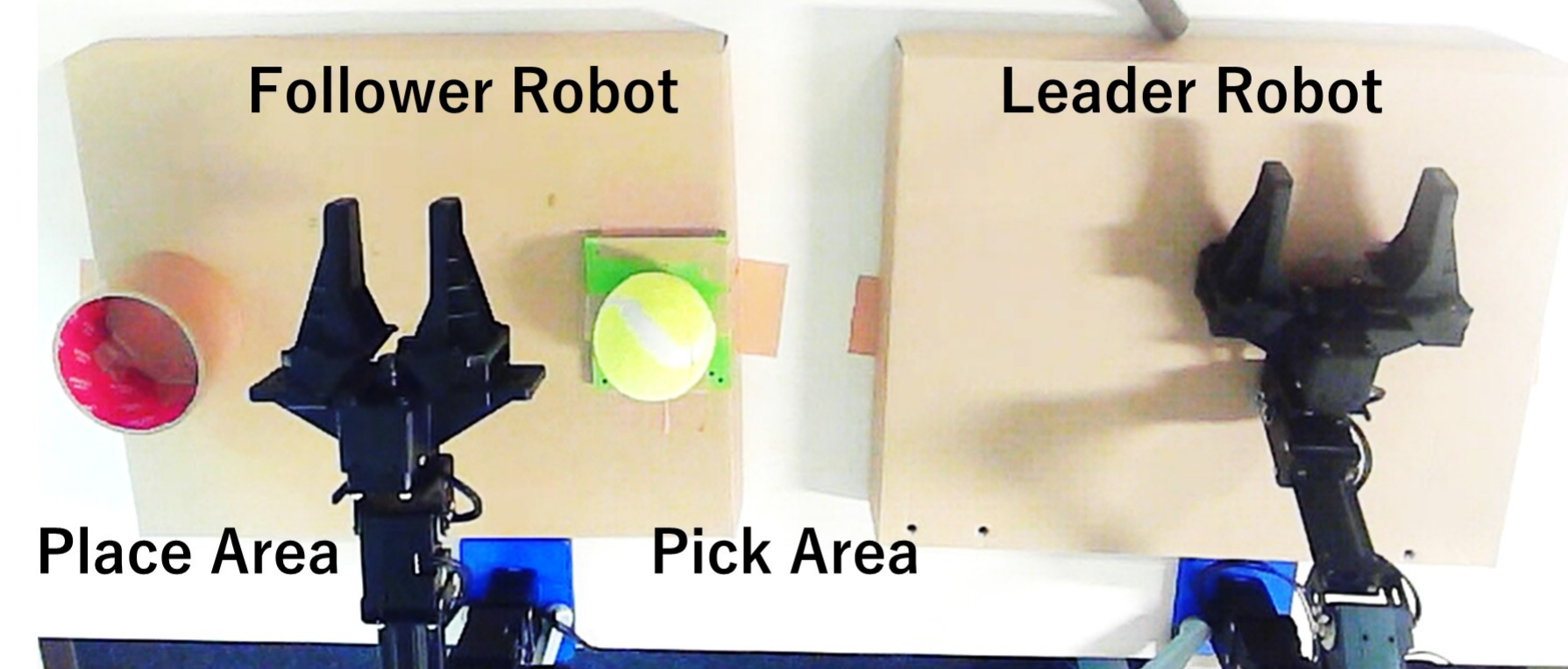}}
      {\begin{center} (a) Robots \end{center}}
    \end{center}
  \end{minipage}
  \begin{minipage}{0.49\hsize}
    \begin{center}
      \scalebox{0.15}{
        \includegraphics{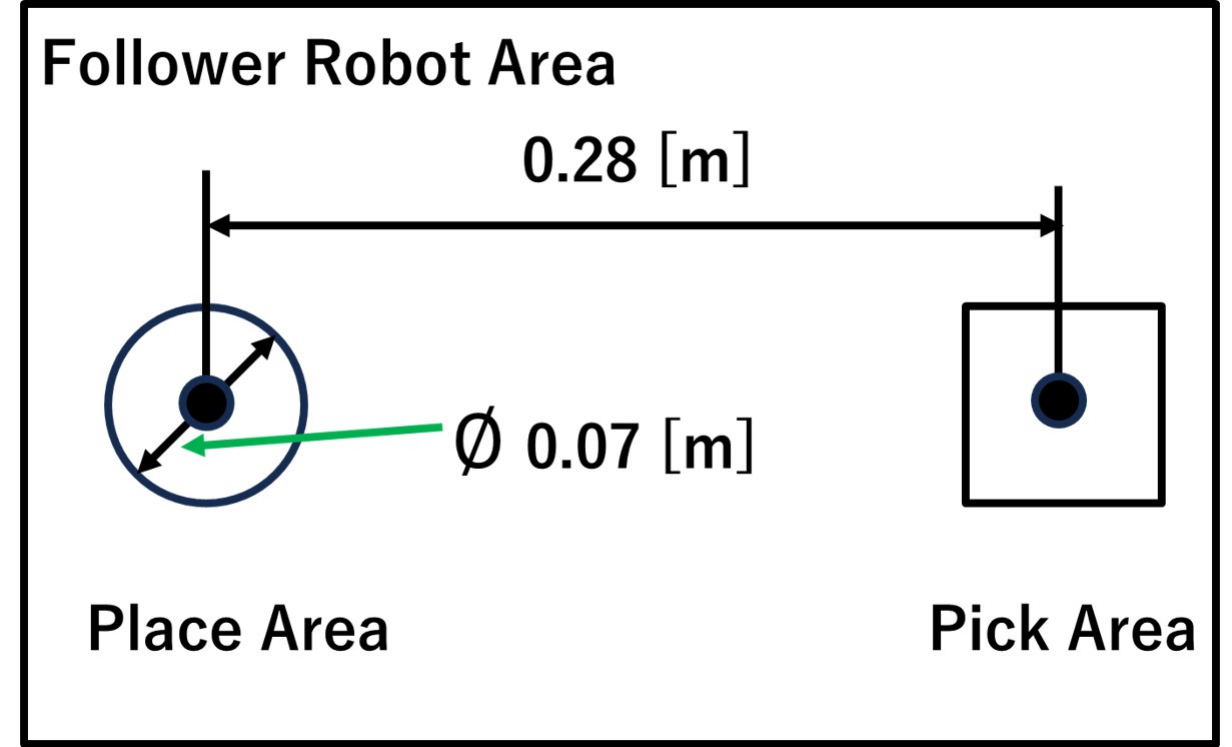}}
      {\begin{center} (b) Image of Environment \end{center}}
    \end{center}
  \end{minipage}
  \caption{Experiment Environment}
  \label{fig:env}
\end{figure}
Fig.~\ref{fig:env} shows the experiment environment.
As shown in Fig.~\ref{fig:env}(a), there are two robots defined as "Follower Robot" and "Leader Robot".
In the conducted experiments, the OpenMANIPULATOR-X robotic arm, manufactured by ROBOTIS, was utilized as shown in Fig.~\ref{fig:OMX}.

Fig.~\ref{fig:env}(b) shows the image of the environment.
It precisely marks out two critical areas: the "Place Area" and the "Pick Area," with explicit spatial dimensions rendered in meters. The representation includes a circle with a diameter of 0.07 meters, denoting the operational radius of the Follower Robot, and a square indicating the Pick Area.
The distance interlinking these two pivotal zones is detailed as 0.28 meters.

\subsection{Task Setting}
\begin{figure}[t]
  \begin{center}
    \scalebox{0.25}{
        \includegraphics{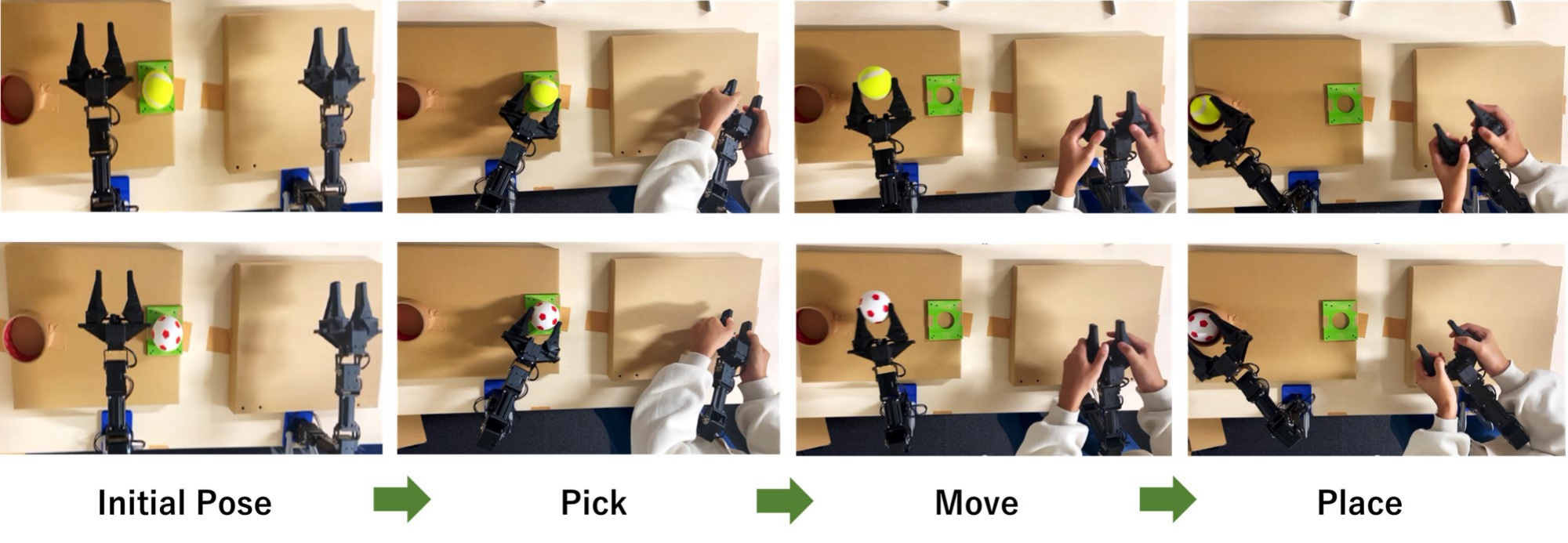}}    
  \caption{Snapshot of Data Collection}
  \label{fig:data}
\end{center}
\end{figure}

\begin{figure}[t]
  \begin{minipage}{0.49\hsize}
    \begin{center}
      \scalebox{0.18}{
        \includegraphics{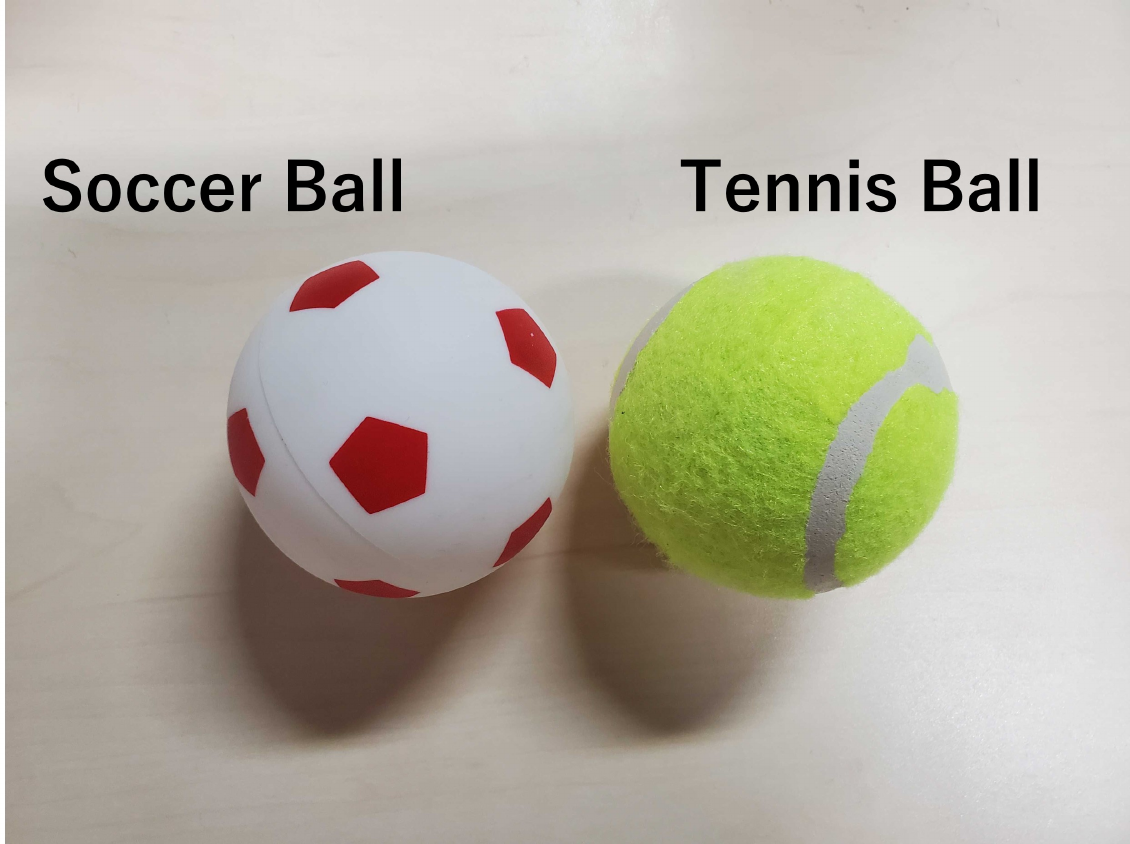}}
      \vspace{-2mm}
      {\begin{center} (a) Trained Objects \end{center}}
    \end{center}
  \end{minipage}
  \begin{minipage}{0.49\hsize}
    \begin{center}
      \scalebox{0.18}{
        \includegraphics{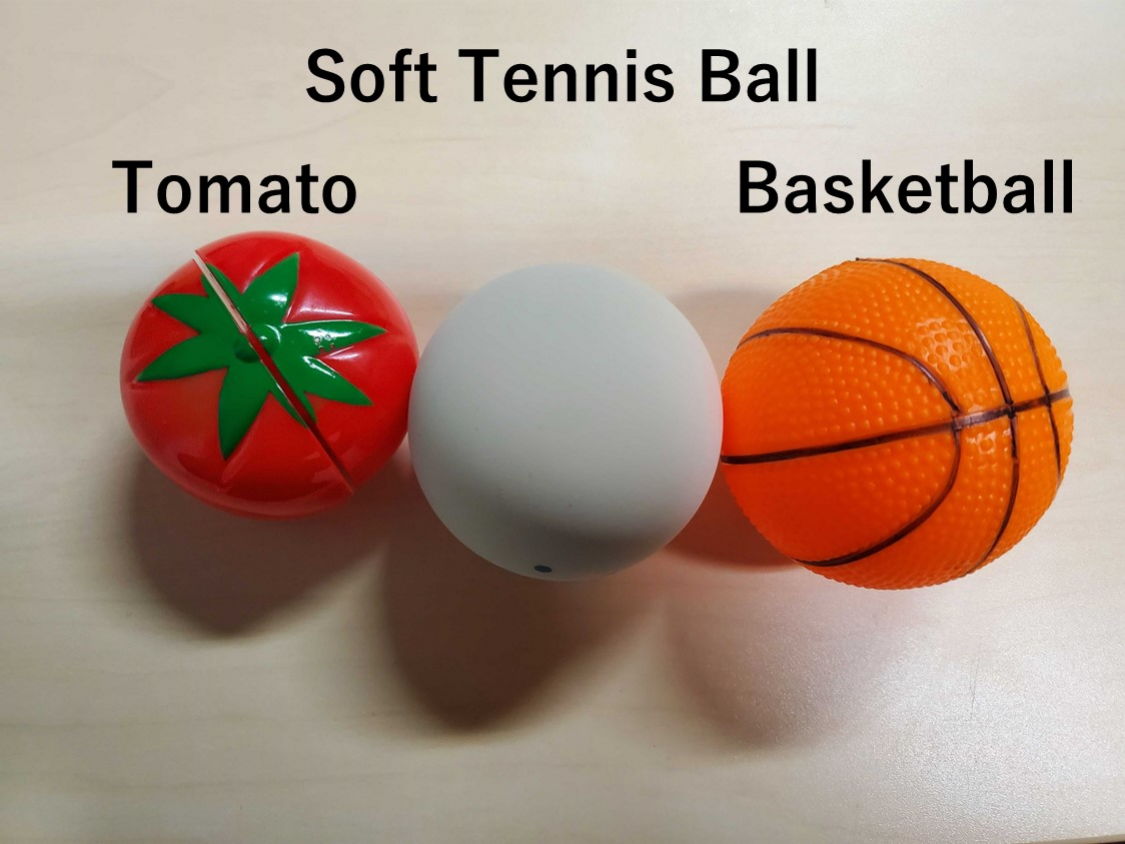}}
      \vspace{-2mm}
      {\begin{center} (b) Untrained Objects \end{center}}
    \end{center}
  \end{minipage}
  \caption{Image of Trained and Untrained Objects}
  \label{fig:objects}
\end{figure}

We conducted a pick-and-place task for objects.
The overall flow of the task is depicted in Fig. \ref{fig:data}.
The objects are initially positioned in the Pick Area.
The task can be divided into three stages.
\begin{enumerate}
    \item \textbf{Pick}: The initial phase of the task, where the robot grasps and lifts the object. This stage requires precise control to securely grip the object.
    
    \item \textbf{Move}: This stage involves transporting the object from the pick location to the designated place location. It demands stable and controlled movement, ensuring that the object remains secure during transit.
    
    \item \textbf{Place}: The final stage, where the robot places the object within the specified area at the place location.
\end{enumerate}
The criteria for success were defined as placing the object within the Place Area without it falling at any point after being grasped.
For experiments, we used a hard tennis ball and a soft rubber soccer ball as the training data, and a very hard object that simulates a tomato, soft tennis ball, and the basketball as untrained objects, as shown in Fig.~\ref{fig:objects} and Table~\ref{tbl:data_obj}.

\subsection{Training Dataset}
\begin{table}[t]
  \caption{Detail of Trained and Untrained Objects}
    \scalebox{0.85}{
\begin{tabular}{cccccc}

\hline
                   & \textbf{Tennis}  & \textbf{Soccer}      & \textbf{Tomato}    & \textbf{Soft Tennis} & \textbf{Basketball}              \\ \hline
Training Data    &  \ding{52}             &  \ding{52}                & \ding{54}    & \ding{54} & \ding{54}         \\
Size & Medium & Medium  & Small & Medium & Large  \\
Hardness & High & Low & Very  High & Very Low & Low \\ 
Weight [g] & 69 & 21 & 14 & 29 & 25 \\ \hline
\end{tabular}
}
\label{tbl:data_obj}
\end{table}
For the tennis ball and soccer ball, the task was performed 25 times each, collecting a total of 50 demonstration dataset using bilateral control, as shown in Fig.~\ref{fig:data}.
In this experiment, we collected data on the robot's angle, angular velocity, and torque at 100 Hz for training.
Analysis of the collected data revealed a range of training values across the 50 demonstrations.
The minimum and maximum recorded timesteps of demonstrations were 889 and 1314 per episode respectively, showcasing the range of dataset sizes within our experiments.
From these 50 demonstrations, we obtained extensive data on the robot's joint angle, velocity, and torque. In this setup, each motor produced data for 54,356 steps, culminating in a total of 1,630,680 data points for training. This data was amassed from two robots, each equipped with five motors, capturing angle, velocity, and torque. The data collection was computed as 54,356 steps × 5 motors × 3 data types × 2 robots.

\subsection{Training Model}
Fig.~\ref{fig:lstm} and~\ref{fig:TE} present the overview of the IL models used in the experiment.

The conventional approach utilizes an LSTM-based model, as depicted in Fig.~\ref{fig:lstm}. The LSTM network is characterized by an input dimension of 15 and hidden units size of 400.
It consists of 6 stacked layers to enhance the model's ability to learn complex patterns.
The final output of the LSTM is fed into a fully connected linear layer with an input size of 400 and an output size of 15, thus matching the dimensionality of the target space.
This model is trained with a learning rate of \(1 \times 10^{-4}\), and a batch size of 32 to ensure adequate convergence.

The proposed method employs a Transformer model, which is structured as follows: The Transformer's encoder comprises 4 layers, each consisting of a multi-head attention mechanism with an output projection from 15 input features to 15 output features.
Each encoder layer also contains two linear transformation steps—the first mapping 15 input features to 240 hidden units, followed by a second mapping the 240 hidden units back to 15 output features.
Dropout with a probability of \( p = 0.1 \) is applied after each of these linear transformations to prevent overfitting.
Layer normalization is applied after each sub-operation with an epsilon value of \( 1 \times 10^{-5} \), to ensure that the transformed representations have a stable distribution.
The final output of the encoder passes through a linear layer with an input and output size of 15, acting as a fully connected feed-forward network.

Both models use mean squared error (MSE) as the loss function and Adam optimizer for training \cite{adam}.
In this experiment, we optimized the inference of both models to be approximately 100 Hz.

\subsection{Experimental Results}
\begin{table}[t]
  \caption{Experimental Results (Total Succes Rate)}
    \scalebox{0.9}{
\begin{tabular}{cccccc}

\hline
                   & \textbf{Tennis}  & \textbf{Soccer}      & \textbf{Tomato}    & \textbf{Soft Tennis} & \textbf{Basketball}              \\ \hline
    LSTM  &  8 / 10 &  5 / 10 & 5 / 10 & 4 / 10 & 5 / 10	\\
    ILBiT(Ours) & 9 / 10 &  9 / 10 & 8 / 10 & 10 / 10 & 7 / 10\\ \hline
\end{tabular}
}
\label{tbl:tab_ex1}
\end{table}

\begin{table*}[t]
  \caption{Experimental Results (Detail Succes Rate)}
    \scalebox{0.85}{
\begin{tabular}{c|ccc|ccc|ccc|ccc|ccc}

\hline
                  & & \textbf{Tennis}& & & \textbf{Soccer} &  &   & \textbf{Tomato}& &   & \textbf{Soft Tennis}& & & \textbf{Basketball} &             \\ 
                  & \textbf{Pick} & \textbf{Move}& \textbf{Place} & \textbf{Pick} & \textbf{Move}& \textbf{Place} & \textbf{Pick} & \textbf{Move}& \textbf{Place} & \textbf{Pick} & \textbf{Move}& \textbf{Place} &  \textbf{Pick} & \textbf{Move}& \textbf{Place}    \\ \hline

    LSTM  &  9 / 10 &  8 / 10 &  8/ 10 & 7 / 10 & 7 / 10 &  5 / 10 &  6 / 10 & 5 / 10 & 5 / 10 & 6 / 10 &  4 / 10 &  4 / 10 & 5 / 10 & 5 / 10 & 5 / 10	\\
    ILBiT(Ours) & 9 / 10 &  9 / 10 & 9 / 10 & 10 / 10 & 10 / 10 & 9 / 10 &  8 / 10 & 8 / 10 & 8 / 10 & 10 / 10 & 10 / 10 &  10 / 10 & 10 / 10 & 10 / 10 & 7 / 10 \\ \hline
\end{tabular}
}
\label{tbl:tab_ex2}
\end{table*}

\begin{figure*}[t]
  \begin{center}
    \scalebox{1}{
     \includegraphics[scale=0.41]{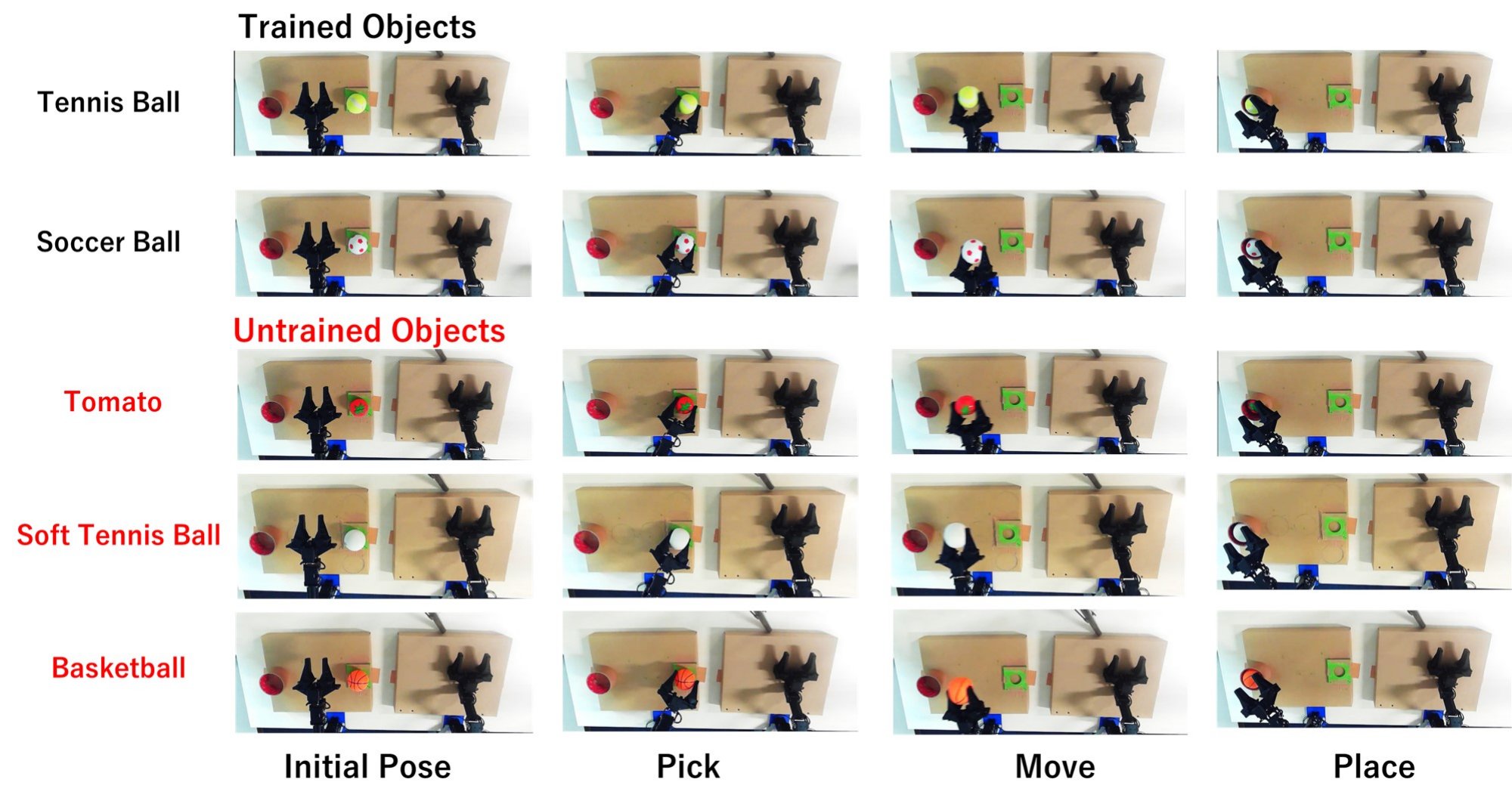}}
  \caption{Experimental Results (Proposed Method)}
  \label{fig:bc}
  \end{center}
\end{figure*}

\begin{figure}[t]
  \begin{center}
    \scalebox{1}{
     \includegraphics[scale=0.23]{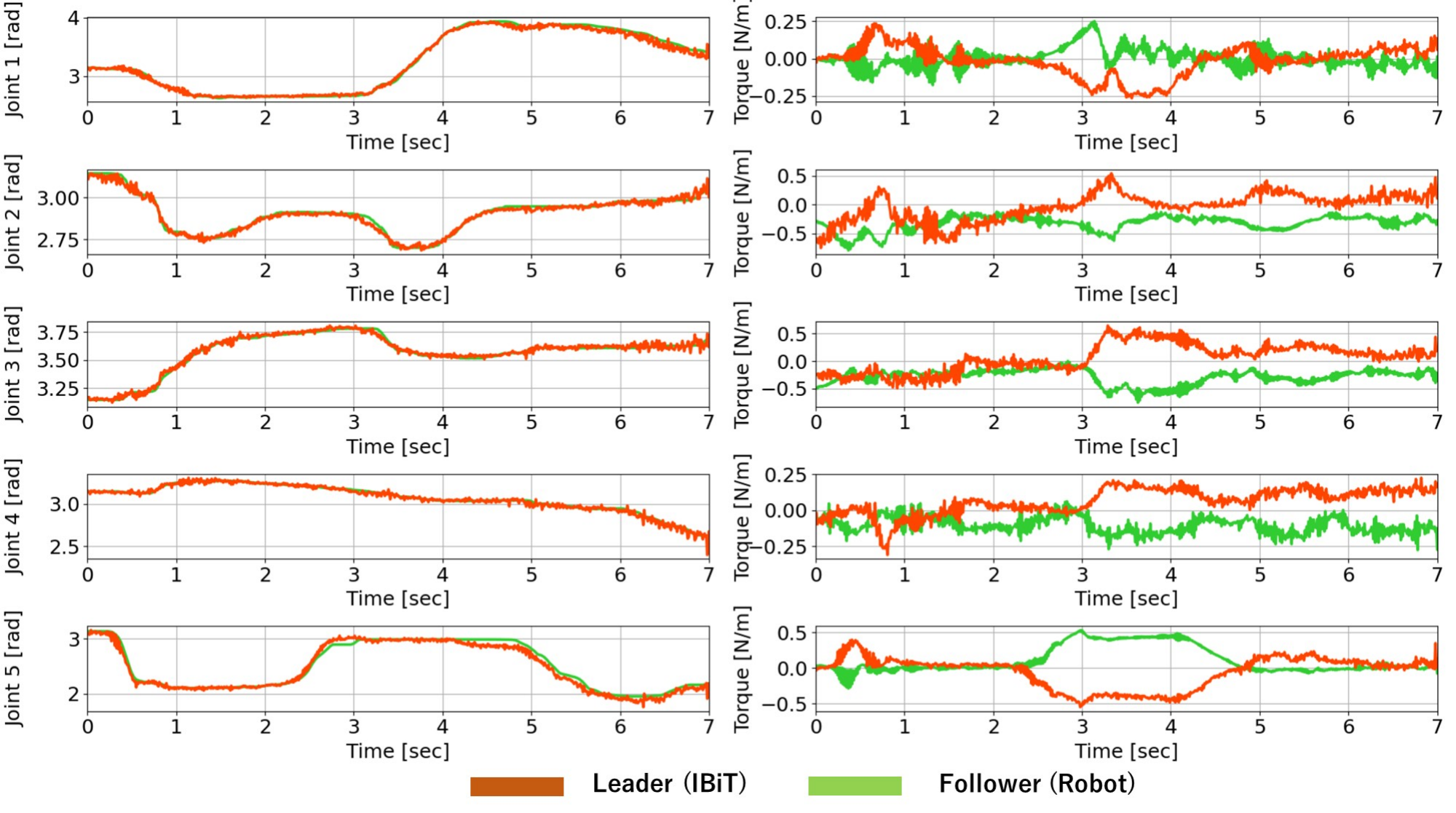}}
  \caption{Results of Joint Angle and Torque in Proposed Method  (Soccer)}
  \label{fig:bc_sc}
  \end{center}
\end{figure}

\begin{figure}[t]
  \begin{center}
    \scalebox{1}{
     \includegraphics[scale=0.23]{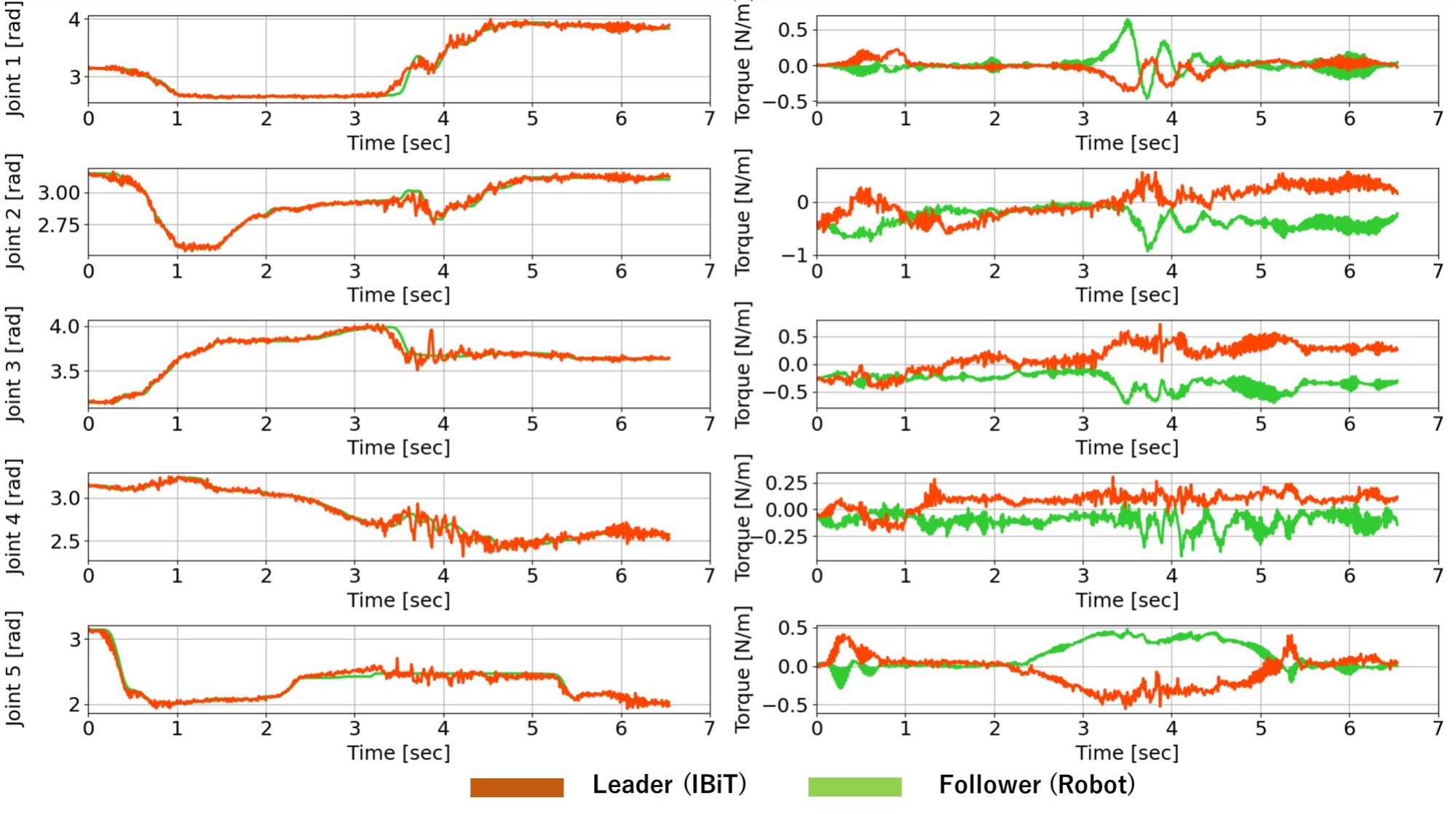}}
  \caption{Results of Joint Angle and Torque in Proposed Method (Tennis)}
  \label{fig:bc_ts}
  \end{center}
\end{figure}

\begin{figure}[t]
  \begin{center}
    \scalebox{1}{
     \includegraphics[scale=0.23]{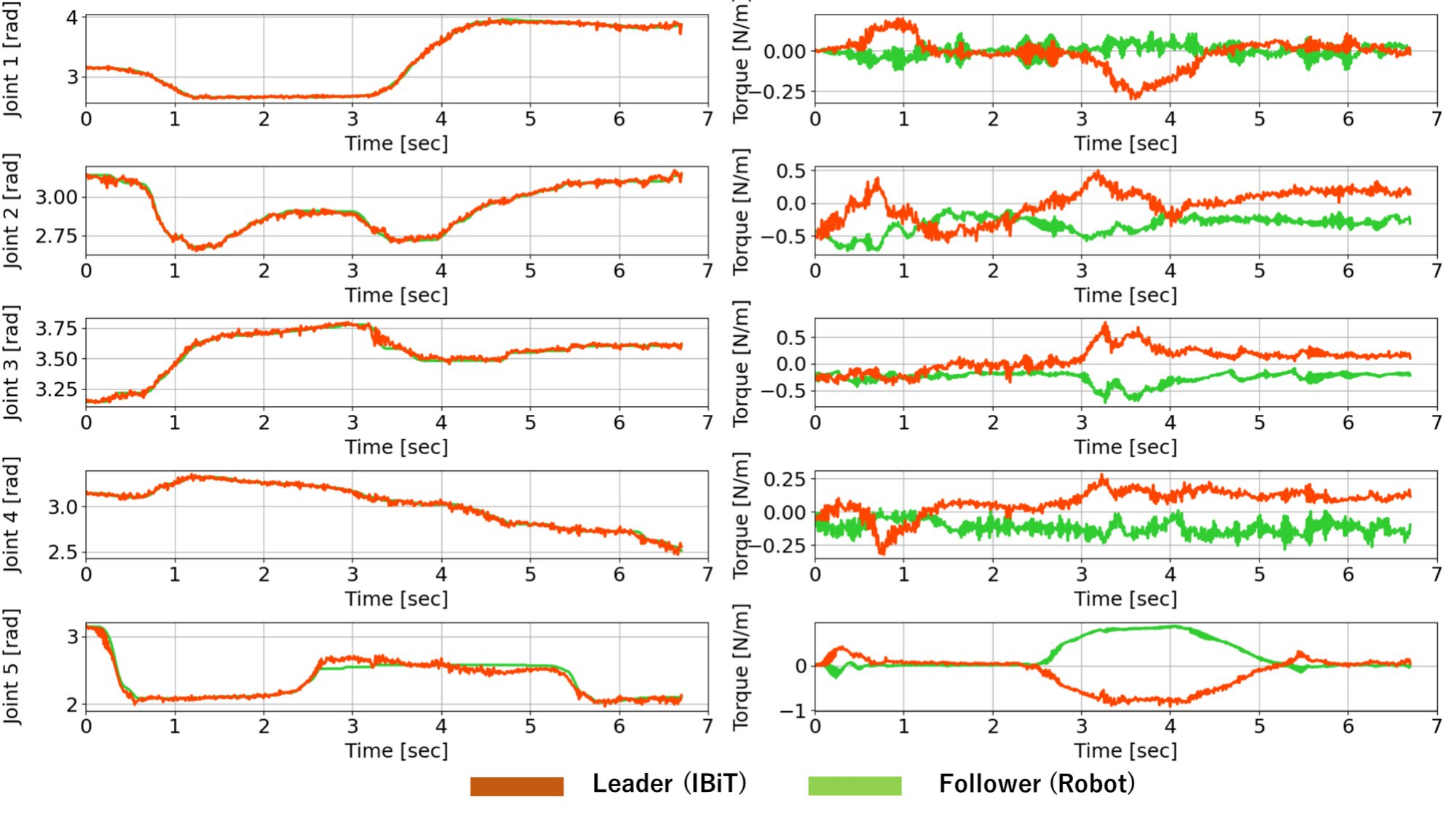}}
  \caption{Results of Joint Angle and Torque in Proposed Method (Tomato)}
  \label{fig:bc_tt}
  \end{center}
\end{figure}

\begin{figure}[t]
  \begin{center}
    \scalebox{1}{
     \includegraphics[scale=0.23]{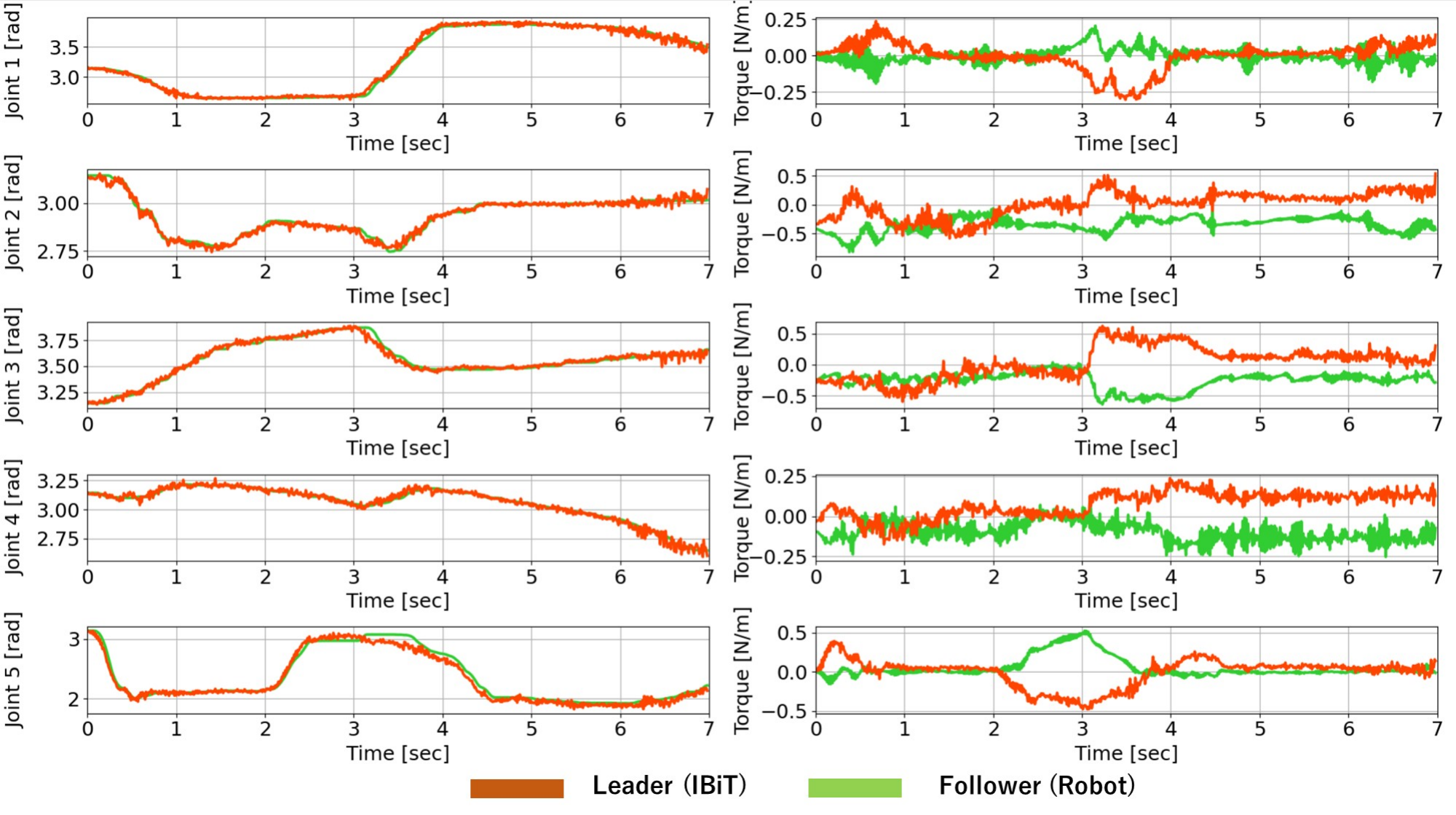}}
  \caption{Results of Joint Angle and Torque in Proposed Method (Soft Tennis)}
  \label{fig:bc_st}
  \end{center}
\end{figure}

\begin{figure}[t]
  \begin{center}
    \scalebox{1}{
     \includegraphics[scale=0.23]{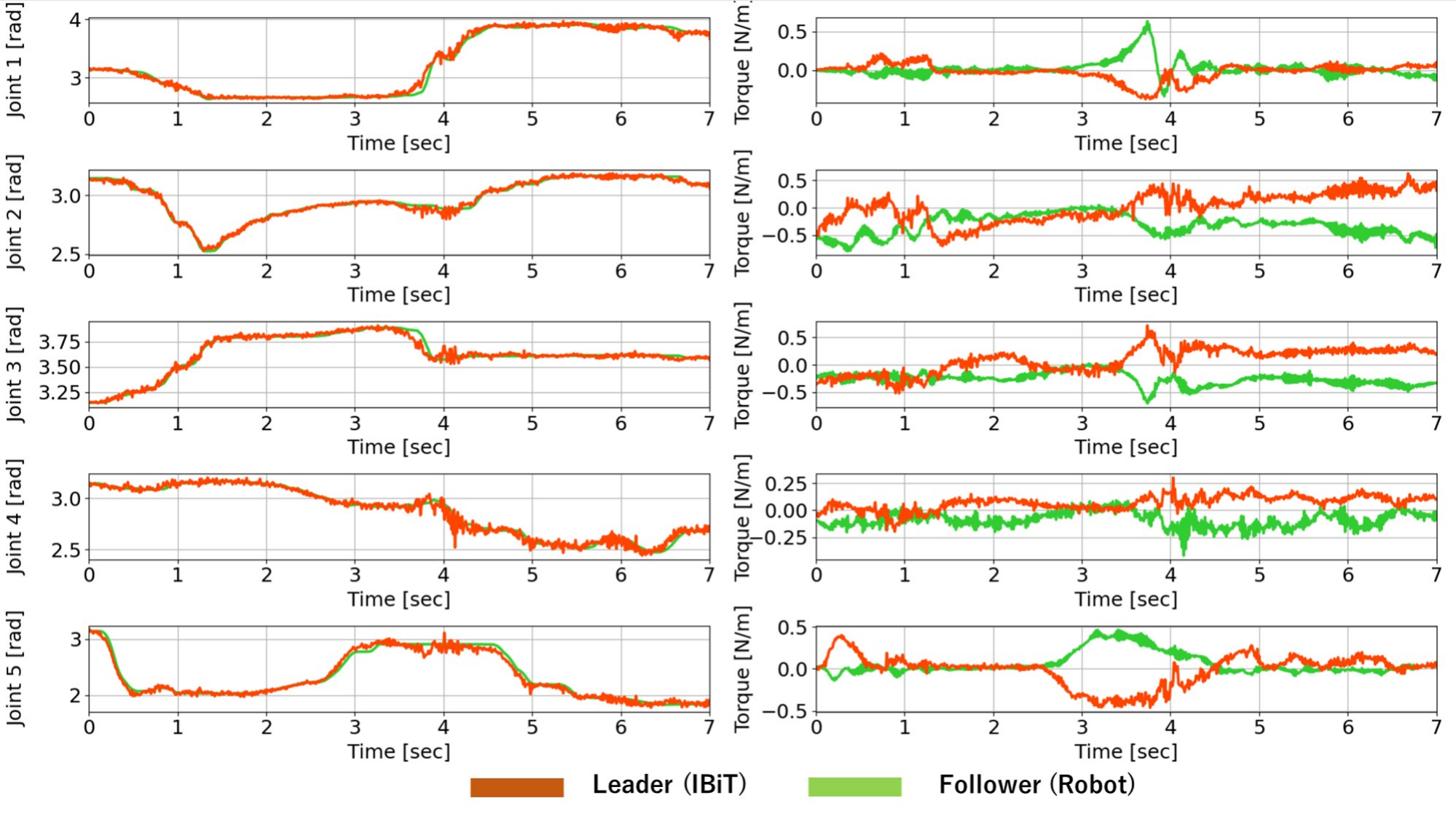}}
  \caption{Results of Joint Angle and Torque in Proposed Method (Basketball)}
  \label{fig:bc_bb}
  \end{center}
\end{figure}

Table~\ref{tbl:tab_ex1} and \ref{tbl:tab_ex2} display the success rates using pretrained models of LSTM (conventional method) and ILBiT, our proposed Transformer-based method, for various objects: tennis ball, soccer ball, simulated tomato, soft tennis ball, and basketball.

Table~\ref{tbl:tab_ex1} reveals that using the LSTM method, the success rates for tennis balls and soccer balls, which were included in the training data, were 80\% and 50\%, respectively.
This illustrates the effectiveness of the conventional method in familiar scenarios.
However, for the untrained objects which are simulated tomato, soft tennis ball and basketball, the success rate dropped to 50\%, 40\%, and 50\%.
Conversely, the Transformer-based ILBiT method demonstrated enhanced performance with success rates of 90\% for both tennis and soccer balls.
For the untrained objects which are the simulated tomato, soft tennis ball and basketball, the success rate was 80\%, 100\%, and 70\% respectively.
Therefore, the proposed method indicated superior adaptability and generalization capabilities.

Further examination in Table~\ref{tbl:tab_ex2} delineates the performance across specific tasks: Pick, Move, and Place.
The LSTM method shows varying success rates, with the highest being 90\% in the Pick task for the tennis ball and the lowest at 40\% in the Place task for the soft tennis ball.
The ILBiT method, however, consistently outperforms LSTM across all tasks and objects.
Notably, it achieves a 100\% success rate in several tasks, including the Pick, Move, and Place tasks for the soft tennis ball.

When combined with the operational data from successful trials using the Transformer model, as illustrated in Fig.~\ref{fig:bc_sc}-\ref{fig:bc_bb}, we observe a clear pattern of effective position tracking and action-reaction relationships. These time-series graphs depict the precision and response of the system over time, highlighting the dynamic interaction between the leader and follower robots during task execution. The coherence of these results with the time-series data underscores the proposed method's potential for real-world applications, where adaptability and precision are paramount. The aggregate of these findings robustly supports the adoption of the Transformer-based approach, showcasing its superior performance in generalizing to new objects and maintaining high success rates across diverse tasks.

The experimental outcomes strongly advocate for the efficacy of the Transformer-based approach, particularly in its ability to generalize to new objects and maintain high success rates across a variety of tasks.

\section{Conclusion}
This paper has proposed the IL based on bilateral control with transformer (ILBiT) for enhancing the capabilities of robotic arms.
This approach intricately combines the robustness of bilateral control with the advanced computational prowess of the Transformer architecture to process position and torque information for complex manipulative tasks.

We meticulously explored the application of the Transformer model within the context of bilateral control.
Through this exploration, ILBiT has been shown to significantly outperform conventional LSTM-based methods in handling diverse dataset.
This advancement facilitates more effective learning and accurate prediction of robotic behavior in time series tasks, thereby improving the efficacy of autonomous manipulation.
The effectiveness of the proposed method was validated through real-world experiments.

In the future, we aim to refine and improve ILBiT.
ILBiT has shown promising results in controlled settings, its robustness in dynamic environments with unpredictable elements remains to be tested. Future research will focus on the integration of real-time adaptability into ILBiT, enabling it to cope with changes such as moving objects and variable lighting conditions.
We also plan to extend our evaluations to a wider range of robotic platforms, including different arm models and sensor setups, to ascertain the generalizability of the ILBiT framework.






\vfill

\end{document}